\documentclass[letterpaper]{article} % DO NOT CHANGE THIS
\usepackage{aaai2027}  % DO NOT CHANGE THIS
\usepackage[hyphens]{url}  % DO NOT CHANGE THIS
\usepackage{graphicx} % DO NOT CHANGE THIS
\usepackage{natbib}  % DO NOT CHANGE THIS AND DO NOT ADD ANY OPTIONS TO IT
\usepackage{caption} % DO NOT CHANGE THIS AND DO NOT ADD ANY OPTIONS TO IT
\usepackage{booktabs}
\newcommand{\kvint}[1]{INT#1}
\newcommand{\CVMA}{\textsc{CVMA}}

\title{Seen, Said, or Forgotten? A Causal Audit of Visual KV Memory Across Dialog Turns}

\author{
    Hong Chen\textsuperscript{\rm 1}\equalcontrib,
    Kang Chen\textsuperscript{\rm 2}\equalcontrib,
    Yuxuan Fan\textsuperscript{\rm 1},
    Bo Wang\textsuperscript{\rm 1},
    Yubo Gao\textsuperscript{\rm 1},
    Yuanlin Chu\textsuperscript{\rm 1},
    Xuming Hu\textsuperscript{\rm 1}\corresponding
}

\affiliations{
    \textsuperscript{\rm 1}The Hong Kong University of Science and Technology (Guangzhou)\\
    \textsuperscript{\rm 2}Wenzhou-Kean University\\
    \{hchen763, yfan546, bwang423, ygao704, ychu763\}@connect.hkust-gz.edu.cn,\\
    chenkang@kean.edu,\\
    xuminghu@hkust-gz.edu.cn
}

\begin{document}

\maketitle

\begin{abstract}
Stateful multimodal assistants encode an image once but may answer questions about it many turns later. Attention-guided visual-KV eviction assumes that evidence irrelevant now will remain dispensable, although future questions are unknown. We ask when a visual fact is actually safe to forget and introduce the \emph{Causal Visual Memory Audit} (\CVMA), a paired single-prefill framework that tests what later answers lose when a visual region, the whole image, or prior assistant text becomes unavailable. On VisDial, current attention can rank future-useful regions worse than random even though a diagnostic marginal-utility control shows substantial selection headroom. Across VisDial and ConvBench, aggregate scores hide accessibility loss when later turns do not need vision; controlled and stock-generated histories reveal a second escape route, in which assistant-text KV replaces image KV for facts already stated but not reliably for unstated facts. In the tested stacks, low current attention provides no safe-forgetting certificate; independently low future dependence and fact-specific verbalization are two conditions under which deletion can appear safe.
\end{abstract}

\section{Introduction}

Multimodal assistants increasingly operate \emph{statefully}: an image is encoded once, then discussed over many turns while its KV cache remains resident. A large body of work reduces that cache by scoring visual entries with current attention and irreversibly evicting low-scoring ones \citep{wan2024look,tu2025vl,wang2026prefixkv}. The choice is efficient, but it makes a consequential identification: what matters now is assumed to be what will matter later.

Consider an image containing a red mug and a clock. The first question asks what is on the table, so the model attends to the mug and evicts the low-scoring clock. Two turns later, the user asks for the time. The clock was \emph{seen}, but it was neither retained nor \emph{said}; its visual evidence is now \emph{forgotten}. Was it ever safe to forget merely because it was irrelevant to the first question?

Prior multi-turn work recognizes that future questions can change which visual content matters \citep{khaki2025sparsevila,wang2026rethinking,liu2026retentivekv}, but aggregate outcomes cannot resolve this example. A small average loss may mean that a selector preserved useful evidence, that later turns never needed vision, or that a needed visual fact had already been verbalized into assistant text. These explanations imply different memory mechanisms, yet ordinary evaluation conflates them. Prior visual-to-text work likewise studies migration within a forward pass \citep{lin2025boosting} or constructs explicit verbalized memory \citep{chatterjee2025memory}, rather than testing when naturally occurring assistant-output KV replaces persistent image KV across dialog turns.

\begin{figure*}[!t]
\centering
\includegraphics[width=0.86\textwidth]{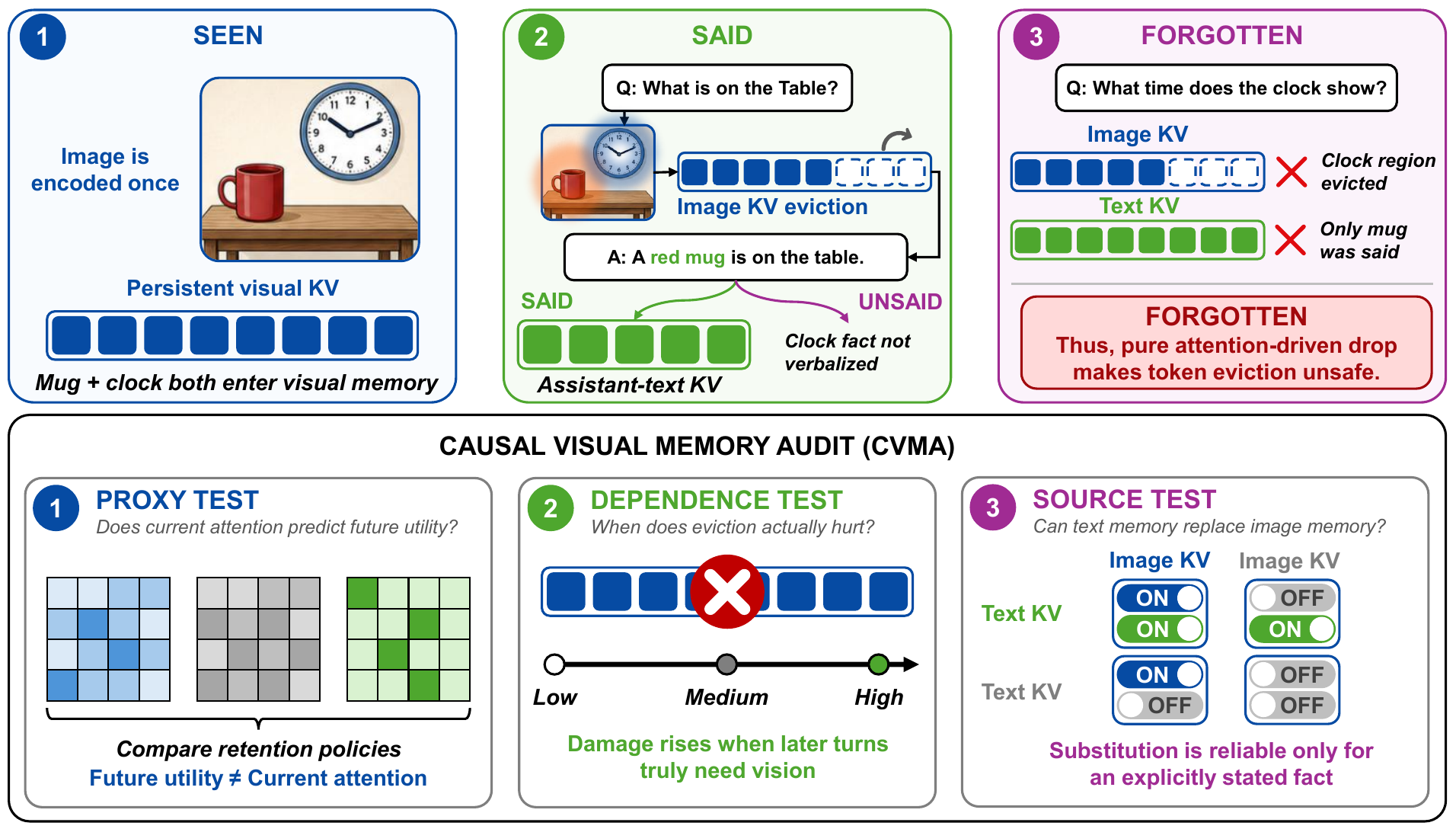}
\caption{\textbf{When is visual memory safe to forget?} The top trajectory exposes the hidden commitment behind current-attention eviction: the clock is seen once, ignored by the current question, left unsaid, and unavailable when a future question needs it. The bottom shows \CVMA's three linked tests of the proposed chain---proxy validity, visual dependence, and fact-specific image--text substitution.}
\label{fig:audit-overview}
\end{figure*}

We turn safe forgetting into a causal question with the \emph{Causal Visual Memory Audit} (\CVMA; Figure~\ref{fig:audit-overview}). \CVMA{} preserves one stateful trajectory---the image is prefetched exactly once---and applies paired interventions within the same dialog. Its components answer three linked questions: does current attention identify future causal utility; when does a wrong eviction decision become harmful; and when can assistant-text KV rescue the missing visual fact? Regional drops, all-image drops, random and marginal-utility retention controls, and image--text factorial interventions make the competing explanations separately testable.

\textbf{In plain terms,} attention measures what helps the answer now, whereas \CVMA{} measures what will be missed later, so a low attention score is not a certificate for forgetting.

Each result motivates the next. Attention is negatively associated with future regional utility and joint retention is worse than random, though a marginal-utility control shows selection still has headroom. This proxy failure should be catastrophic, yet aggregate degradation is often modest because damage concentrates on visually dependent dialogs and later turns while low-dependence cases even benefit from denoising. The remaining route is fact-level: assistant-output KV can replace image KV for facts placed on the answer surface, whereas hidden traces of unstated facts are model-dependent and not reliably answer-discriminative.

Our contributions are methodological and mechanistic, not a new compression policy. (1) \CVMA{} is a reusable analysis framework with causal estimands, state invariants, random/marginal-utility controls, and dependence-conditioned reporting. (2) It identifies attention-guided irreversible retention as a mis-specified proxy for future utility and maps when that error matters across budgets, dialogs, and turns. (3) Controlled, natural-history, and non-Qwen tests locate the fact-specific verbalization boundary between visual and text memory. The practical consequence is concrete: evaluators should expose all-image-drop dependence and unstated-fact failures, while compressor designers should not treat low current attention as sufficient evidence for irreversible deletion. The methodological novelty is the separating intervention system; the empirical novelty is the resulting mechanism chain.

\section{Auditing Safe Forgetting with \CVMA}

\subsection{A stateful paired protocol}
\CVMA{} treats a dialog as one persistent memory trajectory: turn~1 contains the image and every later utterance extends the same cache. Main-text policy results use Qwen2.5-VL-7B-Instruct \citep{bai2025qwen25vltechnicalreport} in BF16, the causal audit also tests Idefics3-8B-Llama3, and a Qwen 3B scale check is in the appendix. Teacher forcing makes answer NLL exactly paired; runtime assertions enforce token-exact extension, immutable visual indices, and complete-source masking (appendix).

VisDial v1.0 validation \citep{das2017visual} provides 2{,}064 ten-turn dialogs; ConvBench \citep{liu2024convbench} provides 547 three-turn long-form dialogs, 546 usable after excluding one documented 9{,}302-token degenerate reference. Full policy comparisons use all 2{,}064/546 dialogs. For the regional audit, we remove 72 earlier pilot/confirmation IDs, shuffle the rest once (Python seed 42), and freeze the first 200 before causal outcomes. The 22-image fact-relay pilot is never pooled with its untouched 80-image confirmation, and the natural-history follow-up is labeled external validity, not a second confirmation.

This evidence spans two model families and two contrasting dialog regimes. We treat agreement across them as replication within the tested stacks, not as a universal claim over VLM architectures, benchmarks, or compression policies.

The outcome $L_{d,t}(Z)$ is mean cross-entropy over the reference answer's content tokens under intervention $Z$. Regional utility and all-image dependence use the future horizon $H=\{2,\ldots,T_d\}$ after the turn-1 anchor---matching the audited decision to compress once before any follow-up is known---while full-policy results average over all turns. ``Drop'' makes the selected cache positions unreadable, leaving cache length, positions, and RoPE indices unchanged.

Current attention is also fixed operationally. For first-turn question positions $Q$, visual positions $V$, layer $\ell$, and head $h$, we compute
\begin{equation}
 a_i=\frac{1}{N_\ell}\sum_\ell
 \frac{\frac{1}{N_h|Q|}\sum_{h,q\in Q} A_{\ell hqi}}
 {\sum_{j\in V}\frac{1}{N_h|Q|}\sum_{h,q\in Q} A_{\ell hqj}},
 \qquad i\in V,
\label{eq:attention-score}
\end{equation}
where $N_\ell$ and $N_h$ are the layer and head counts. We then average $a_i$ over the visual tokens assigned to a spatial region. Thus the proxy test uses the same first-turn question, head averaging, per-layer visual-mass normalization, and layer averaging in both model stacks; no future token enters the ranking.

\subsection{Four interventions that separate the explanations}
Let
\begin{equation}
\begin{array}{c}
\bar L_d(Z)=\sum_{t\in H}w_tL_{d,t}(Z),\\[-1pt]
w_t=|H|^{-1},\quad H=\{2,\ldots,T_d\}.
\end{array}
\label{eq:future-loss}
\end{equation}
\emph{Is region $r$ useful later?} Regional future utility answers by removing it:
\begin{equation}
U_{d,r}=\bar L_d(-r)-\bar L_d(\mathrm{full}).
\label{eq:future-utility}
\end{equation}
This is the causal target that current attention is supposed to predict. \emph{Would the future dialog need image KV at all?} Visual dependence instead removes the whole image span:
\begin{equation}
D_d=\bar L_d(-\mathrm{image})-\bar L_d(\mathrm{full}).
\label{eq:visual-dependence}
\end{equation}
We report $D_d$ continuously and by quantile; ratios with $D_d$ are prohibited because near-zero denominators are unstable. \emph{Is attention the problem, or is selection intrinsically hopeless?} At retention budget $q$, proxy regret compares paired rankings,
\begin{equation}
R_{d,q}^{p-c}=\bar L_d(\mathrm{retain}(p,q))-\bar L_d(\mathrm{retain}(c,q)).
\label{eq:proxy-regret}
\end{equation}
Seeded random tests practical misranking; a same-dialog \emph{marginal-utility control} ranks regions by single-drop $U_{d,r}$. It witnesses selectable signal but is neither deployable nor a joint optimum when regions interact (figures abbreviate it as ``oracle'').

\begin{figure*}[!t]
\centering
\includegraphics[width=0.86\textwidth]{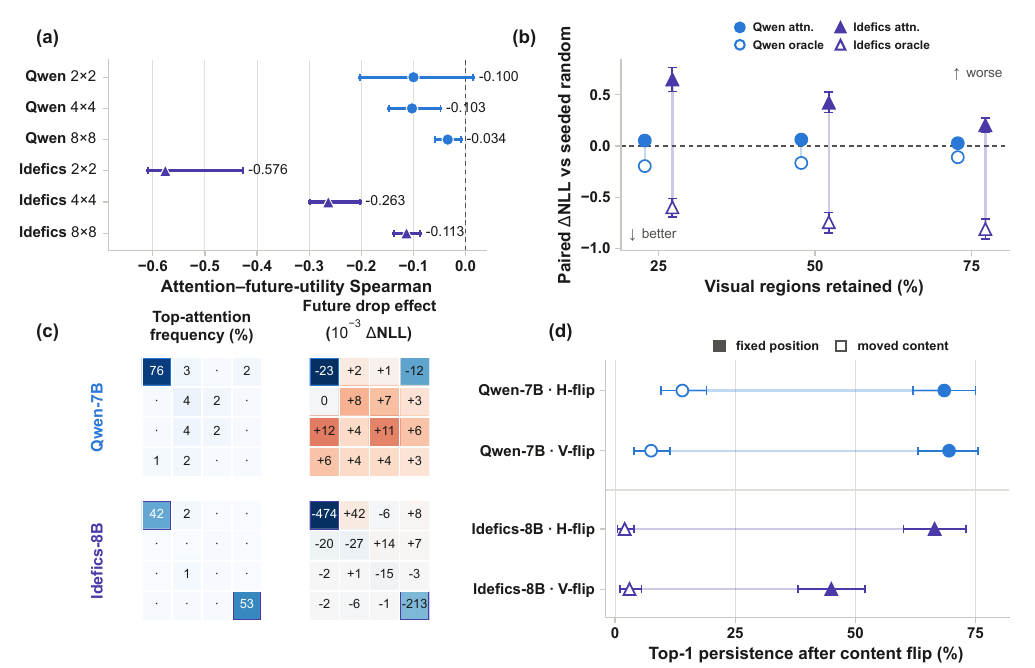}
\caption{\textbf{Proxy-validity tests for current attention.} (a) Mean per-dialog attention--utility Spearman (200 dialogs). (b) Joint-retention $\Delta$NLL at a matched $4{\times}4$ grid for attention, seeded random, and the marginal-utility control (``Oracle''). (c) Top-attention frequency and future-drop effect by cache position, endpoints outlined (percentages and $10^{-3}$ NLL). (d) Top-1 persistence after content flips, at the fixed position versus the moved content. Intervals are 95\% hierarchical (a) or dialog-bootstrap (b,d).}
\label{fig:proxy-validity}
\end{figure*}

Finally, \emph{has a needed visual fact moved into text?} At boundary $b$, let $V$ be image KV and $A_{\le b}$ completed assistant-output KV. For the immediate next answer, their factorial interaction is
\begin{equation}
\begin{array}{rcl}
I_{d,b}&=&L_{d,b+1}(-V,-A_{\le b})-L_{d,b+1}(-V)\\
&&{}-L_{d,b+1}(-A_{\le b})+L_{d,b+1}(\mathrm{full}).
\end{array}
\label{eq:factorial-interaction}
\end{equation}
Positive $I$ indicates redundancy or substitution on the NLL scale; it does \emph{not} by itself identify what moved. The controlled fact experiment below therefore asks whether the target fact itself reached assistant text.

\subsection{Identification assumptions and invariants}
The estimands are causal for the frozen, teacher-forced trajectory under four conditions---intervention consistency, no interference between separately replayed branches, identical pre-intervention state within a pair, and no intervention-induced change to future history tokens (enforced by teacher forcing). This bounds claims to the declared trajectory rather than to free generation. The appendix states the full conditions, the exact cache-read procedure, and the executable audit manifest.

\subsection{Compression controls}
We compare five selectors under one Qwen2.5-VL stateful runtime at exact visual budgets---static first-turn top-$k$, LOOK-M pivotal merging \citep{wan2024look}, PrefixKV layer-adaptive retention \citep{wang2026prefixkv}, SnapKV observation-window pooling \citep{li2024snapkv}, and NACL proxy-plus-random retention \citep{chen2024nacl}---against a non-deletion coverage control that applies groupwise affine fake quantization with per-channel keys as in KIVI \citep{liu2024kivi} and an optional Walsh--Hadamard rotation (accounted bits include scale/zero-point metadata: \kvint{2} at group 32 costs 3 bits/element). Because low-bit coverage is prior art \citep{liu2024kivi,yang2024no,han2025calibquant}, no method or serving-latency claim is attached to it. Released VisionZip \citep{yang2025visionzip} removes visual embeddings before LM prefill, so we compare it only with a paired in-stack full ceiling. The appendix lists each transplant's unavoidable differences; every policy-level comparison uses all five selectors at common budgets.

\section{Attention Is Not Future Utility}

We first ask whether attention under the current question identifies regions whose removal causes future loss; Figure~\ref{fig:proxy-validity}a,b tests this marginally then jointly, and c,d diagnose the spatial contamination behind the failure. On 200 frozen Qwen-7B dialogs, $4{\times}4$ regional attention--future-utility Spearman is $-0.103$ ($[-0.147,-0.048]$; Figure~\ref{fig:proxy-validity}a). It remains negative but smaller at $8{\times}8$ ($-0.034$, $[-0.058,-0.009]$); the coarsest $2{\times}2$ grid is directionally consistent but inconclusive.

Because regional effects may interact, we also retain regions jointly at 25/50/75\%. At the frozen primary $4{\times}4$ grain, attention is worse than seeded random by $+0.053/+0.062/+0.027$ NLL (Figure~\ref{fig:proxy-validity}b); at $8{\times}8$, the gaps are $+0.092/+0.055/+0.016$, again with all CIs excluding zero. Nor is the future intrinsically unselectable here: the $4{\times}4$ same-dialog marginal-utility control beats random by $-0.197/-0.166/-0.110$---a headroom witness rather than an upper bound, since it ranks single-region effects instead of solving the combinatorial problem. Evidence is thus bounded to the medium and fine grains, strongest at the primary $4{\times}4$, not the inconclusive $2{\times}2$, which Idefics probes below. Within the supported cells the problem is proxy misalignment despite available headroom, not selection in principle.

Nor is the result an artifact of the layer aggregation in Equation~\ref{eq:attention-score}: last-four, last-layer, and mass-weighted variants all keep the correlation negative ($-0.084$ to $-0.111$) and joint regret worse than random (Figure~\ref{fig:stress-tests}a,b; Appendix Table~A4), ruling out three immediate aggregation rescues.

The distinction survives a different model family across all three grains: with bit-exact cross-grid controls, Idefics attention--utility correlations are $-0.576/-0.263/-0.113$ at $2{\times}2/4{\times}4/8{\times}8$ (all CIs exclude zero), attention loses to random while the marginal-utility control wins at every grain and budget, and a caption-removal diagnostic rules out VisDial's supplied caption (Figure~\ref{fig:stress-tests}c; appendix gives the per-grain values). Dropping the whole image nevertheless improves Idefics mean NLL, so we replicate the proxy-failure topology, not the sign of aggregate dependence, and treat the latter as a model--task interference boundary.

Finally, the failure is spatially diagnostic (Figure~\ref{fig:proxy-validity}c,d): top attention concentrates at fixed cache endpoints---the first region for Qwen (152/200) and a sequence endpoint for Idefics (190/200)---yet dropping those hotspots improves future NLL, and image-flip controls show the top-1 hotspot follows cache position rather than content. Attention is thus not content-free, but its top-ranked visual position is position-contaminated (appendix).

\section{Eviction Fails Where and When Vision Matters}

Proxy invalidity permits information loss, but not everywhere: a wrong eviction matters only when a later turn needs the discarded evidence. This can explain why attention-guided deletion looks safer in aggregate than its proxy validity suggests.

\subsection{Wrong eviction matters only when vision matters}
\begin{figure*}[t]
\centering
\includegraphics[width=0.82\textwidth]{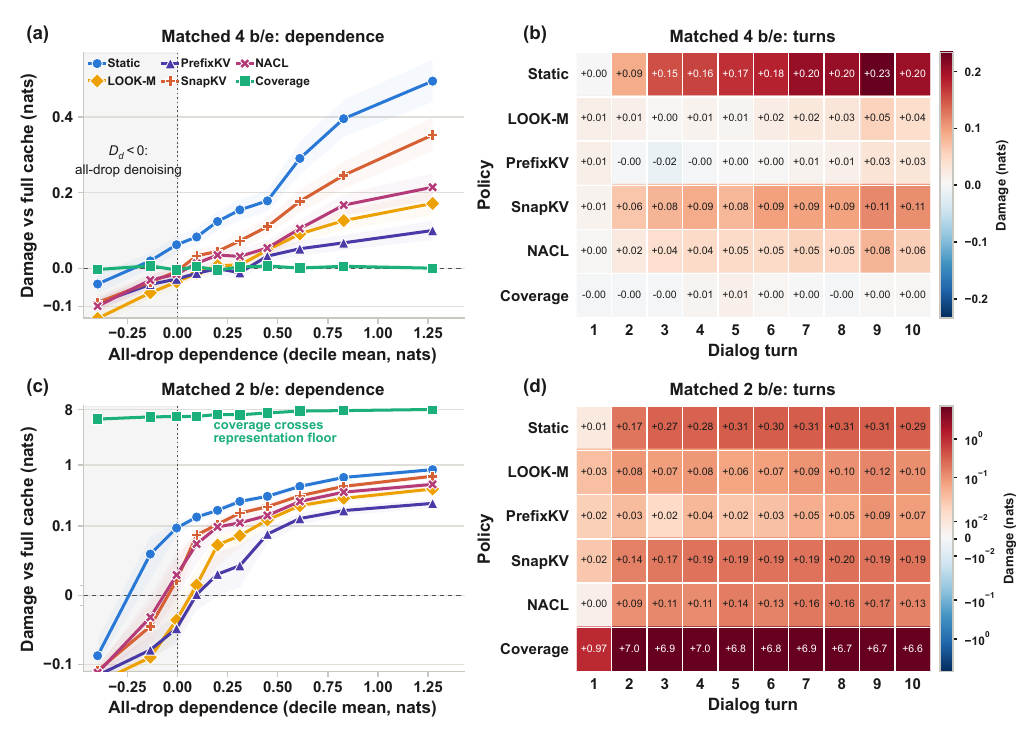}
\caption{\textbf{Dependence- and turn-conditioned damage at exact common budgets.} Full VisDial is binned into ten equal-frequency deciles by an independent all-image drop. (a,b) Damage versus dependence decile and dialog turn at four accounted bits (five 25\% selectors and matched coverage); (c,d) the same at two bits (12.5\% selectors and the one-bit coverage endpoint). The lower row uses symmetric-log scaling with 95\% bootstrap bands.}
\label{fig:floor-drift}
\end{figure*}

\CVMA{} measures this effect modifier independently by dropping all image KV before evaluating the future horizon. Across all 2{,}064 VisDial dialogs, mean dependence is $+0.322$ nats, yet 25.3\% have negative $D_d$. All ten selector--budget cells have positive damage--dependence association (Spearman $0.399$--$0.682$). At the exact four-accounted-bit comparison, coverage has no detectable coupling ($0.012$ $[-0.032,0.055]$); at two bits, its representation itself collapses and dependence coupling reappears ($0.382$ $[0.343,0.419]$). Figure~\ref{fig:floor-drift} keeps both regimes rather than silently dropping the failed endpoint.

Quartiles expose what the mean mixes together. In Q1 ($D_d=-0.222$), LOOK-M/PrefixKV improve NLL by removing distractors; in Q4 ($D_d=+0.971$), static 25/12.5\% retention costs $+0.417/+0.676$ while exact-four-bit coverage costs $+0.004$ (CI crosses zero), and exact-two-bit coverage costs $+7.903$ even in a non-deletion cache. Aggregate evaluation thus offsets irreversible loss on image-dependent dialogs with denoising on dialogs that never needed image KV. Crucially, $D_d$ is a paired within-session intervention tied to the same future outcome, not a benchmark-level necessity label.

Full ConvBench reproduces the topology (association $0.541$--$0.793$; each cell improves Q1 and degrades Q4, while exact-four-bit coverage stays flat at $\rho=-0.009$), so the floor is not specific to VisDial's short answers.

\begin{table}[!t]
\centering
\small
\setlength{\tabcolsep}{3.8pt}
\begin{tabular}{@{}l r r r@{}}
\toprule
Policy & visual b/e & VisDial & ConvBench \\
\midrule
\multicolumn{4}{@{}l}{\emph{Reference}} \\
Full cache (BF16) & 16 & 0.000 & 0.000 \\
\addlinespace[2pt]
\multicolumn{4}{@{}l}{\emph{Attention-guided selection / merging}} \\
Static top-$k$ 25\% & 4 & +0.159 & +0.030 \\
Static top-$k$ 12.5\% & 2 & +0.257 & +0.056 \\
\addlinespace[1pt]
LOOK-M merge 25\% & 4 & +0.021 & +0.018 \\
LOOK-M merge 12.5\% & 2 & +0.080 & +0.036 \\
\addlinespace[1pt]
PrefixKV 25\% & 4 & +0.008 & +0.011 \\
PrefixKV 12.5\% & 2 & +0.040 & +0.025 \\
\addlinespace[1pt]
SnapKV 25\% & 4 & +0.082 & +0.029 \\
SnapKV 12.5\% & 2 & +0.165 & +0.051 \\
\addlinespace[2pt]
\multicolumn{4}{@{}l}{\emph{Attention + randomized retention}} \\
NACL 25\% & 4 & +0.043 & +0.018 \\
NACL 12.5\% & 2 & +0.123 & +0.038 \\
\addlinespace[2pt]
\multicolumn{4}{@{}l}{\emph{Non-deletion coverage}} \\
Coverage INT8 & 9 & 0.000 & 0.000 \\
Coverage INT4 & 5 & +0.001 & 0.000 \\
Coverage INT3 & 4 & +0.002 & 0.000 \\
Coverage INT2+WHT & 3 & +0.008 & +0.002 \\
Coverage INT1+WHT & 2 & +6.229 & +4.974 \\
\bottomrule
\end{tabular}

\caption{Full-dataset Qwen-7B stateful controls (teacher-forced $\Delta$NLL vs. full cache, uniformly averaged over all benchmark turns; lower is better) on VisDial (2{,}064 dialogs) and ConvBench (546). Visual b/e is $16\times$ the BF16 retention fraction for logical selectors (index/kernel overhead excluded) and includes scale/zero-point metadata for coverage; it is not a serving-memory claim. Paired 95\% CIs appear in Appendix Table~A3.}
\label{tab:main}
\end{table}

The effect modification also crosses model families: on the 200 shared dialogs, damage--dependence Spearman is $0.443$ for Qwen and $0.344$ for Idefics, whose damage rises from $+0.060$ in Q1 to $+0.425$ in Q4 despite a negative mean $D_d$. The cross-family invariant is thus ordering---more image-dependent dialogs are more vulnerable---not a universal zero point (Appendix Table~A2).

\subsection{The floor is semantic, not a cache-size artifact}
The association is semantic, not a cache-size artifact: dependence has Spearman $-0.033/-0.126$ with visual-token count on VisDial/ConvBench, yet tracks question type (highest for count/color, lowest for yes/no and reasoning; appendix). Answer length correlates negatively, but length alone cannot tell whether a fact moved into text; the factorial tests below can.

\subsection{Turn-wise failure survives stronger selection}
If the failure comes from an irreversible early decision confronted by changing future queries, it should also emerge over turns. On full VisDial, static 12.5\% damage grows from $+0.008$ at turn~1 to $+0.289$ at turn~10. LOOK-M, PrefixKV, SnapKV, and NACL reduce the level to different degrees but all show larger later-turn than anchor-turn damage at both exact budgets (Figure~\ref{fig:floor-drift}b,d). Viable four-bit coverage stays near zero; failed two-bit coverage is already $+0.97$ at turn~1 and exceeds $+6.6$ thereafter, a qualitatively different representation-error signature. ConvBench is front-loaded at its perception turn, so the selector profile follows task structure rather than a universal monotone law.

Table~\ref{tab:main} gives the fixed-7B policy ladder on both benchmarks, adding source-derived SnapKV and NACL at both budgets. NACL edges SnapKV on aggregate VisDial ($+0.043$ versus $+0.082$ at 25\%), but this does not isolate NACL's randomized component, and neither closes the gap to full cache (PrefixKV $+0.008$; exact-four-bit coverage about zero). More importantly, both added selectors reproduce the failure topology: at 12.5\%, SnapKV/NACL move from $-0.061/-0.062$ in VisDial dependence Q1 to $+0.501/+0.388$ in Q4, and ConvBench reverses likewise. Non-deletion coverage stays closest to full cache only above its precision floor, and even that is not scale-invariant---at 3B, PrefixKV-25\% costs $+0.001$ versus $+0.007$ for coverage. Our claim concerns accessibility topology, not a universal aggregate winner (Appendix Table~A1 gives the full 3B/7B ladder).

\begin{figure}[t]
\centering
\includegraphics[width=\columnwidth]{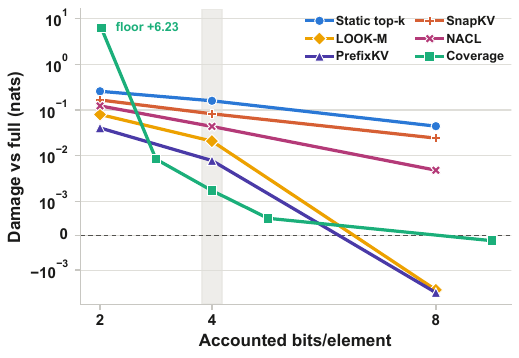}
\caption{\textbf{Matched-storage damage across the accounted-bit ladder.} The aligned full-VisDial sweep gives every selector 50/25/12.5\% retention and covers coverage's full measured precision range; the shaded slice marks four-bit accounting. The symmetric-log axis has a linear zone of $\pm10^{-3}$ NLL and retains the failed two-bit endpoint.}
\label{fig:pareto}
\end{figure}

The aligned sweep also exposes a representation boundary the selector comparison cannot show. Coverage stays within $+0.008$ NLL of full cache from nine down to three accounted bits, but collapses by $+6.229$ at two bits (one-bit codes plus affine metadata, after WHT) and by $+4.974$ on full ConvBench, versus about zero at four bits (Table~\ref{tab:main}). Persistent addressability is thus not sufficient at arbitrary precision: coverage is a valid control only while its representation stays decodable. Figure~\ref{fig:pareto} reports the failed VisDial endpoint rather than truncating the curve.

\begin{figure*}[!t]
\centering
\includegraphics[width=0.82\textwidth]{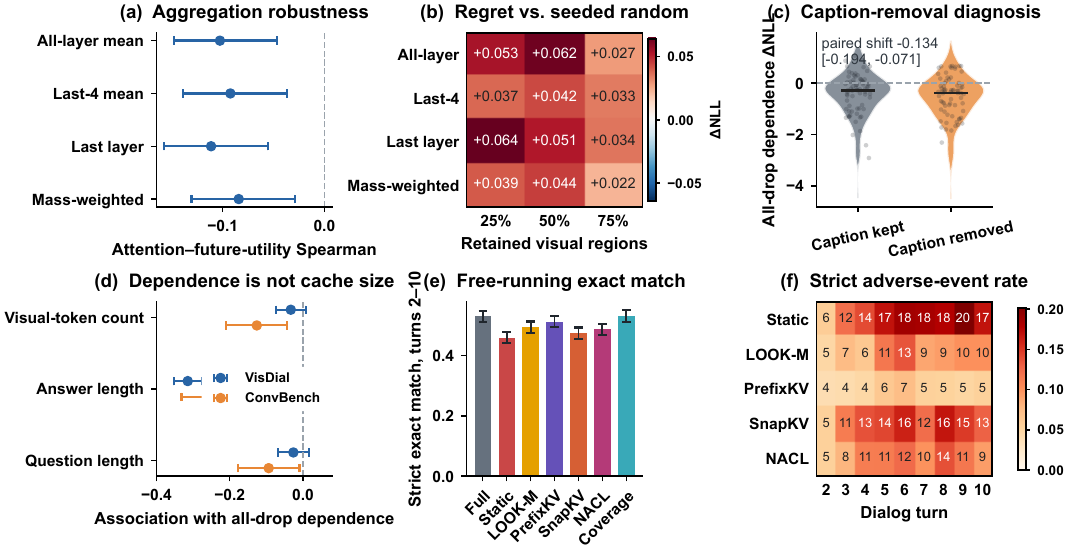}
\caption{\textbf{Robustness checks and free-running answer quality.} (a) Attention--utility correlation under four aggregation rules. (b) Joint-retention $\Delta$NLL versus seeded random; positive is worse. (c) Idefics all-image-drop dependence with VisDial's caption kept versus removed. (d) All-drop dependence versus visual-token count, answer length, and question length. (e) Strict exact match on turns 2--10 of all 516 dependence-Q4 dialogs at 25\%/four-bit. (f) Among cells where full and coverage match the reference, the percentage each selector does not, by turn. Error bars 95\% dialog-bootstrap; details in Appendix Tables~A4, A5.}
\label{fig:stress-tests}
\end{figure*}

\subsection{From hidden NLL drift to visible failure}
Teacher-forced NLL is primary because it keeps every paired future trajectory fixed; a sealed audit of the 15 highest-dependence dialogs (14 usable after a chat-template guard) bridges it to observable errors. The dominant failure is late-turn \emph{hallucinated absence} after peripheral evidence is deleted, while full cache, \kvint{2}+WHT, and the reference agree and no cell reverses that relation (Appendix Table~A6 gives counts and verbatim cases).

We then freeze the complete high-dependence quartile---516 dialogs and 4{,}644 later-turn cells---and generate all ten turns under Full, all five 25\% selectors, and matched four-bit coverage. Conservative strict exact match on turns 2--10 is 53.1\% for full cache and 53.3\% for coverage, versus 46.0--51.4\% for the selectors (Figure~\ref{fig:stress-tests}e). Every paired selector gap is negative with its 95\% CI below zero, ranging from $-7.08$ points $[-7.97,-6.22]$ for static to $-1.68$ $[-2.33,-1.01]$ for PrefixKV. Under the stricter directional anchor---full and coverage both correct---each selector produces more adverse than strict reverse cells: 126--377 versus 47--54, affecting 20.7--52.5\% of dialogs (Figure~\ref{fig:stress-tests}f; Appendix Table~A5). Thus the visible effect is neither confined to 14 hand-inspected cases nor to one weak selector, while the original sealed audit supplies their semantic interpretation.

On all 2{,}064 VisDial dialogs, official-formula dense NDCG is $0.557$ full, $0.534/0.515$ at 25/12.5\% static retention, and $0.557/0.554$ for \kvint{4}/\kvint{2}+WHT. Coverage beats matched retention by $+0.0233$ $[0.0184,0.0282]$ and $+0.0390$ $[0.0327,0.0455]$. This is a stateful likelihood result, not a leaderboard submission.

ConvBench's benchmark-native check (all seven policies, one deployment, 546 dialogs) agrees: only the two most aggressive selectors, static and SnapKV, lose first-turn preference to full ($+2.2$ $[0.0,4.6]$ and $+2.9$ $[0.9,5.1]$ on both frozen A/B sides), while LOOK-M, PrefixKV, NACL, and coverage include zero (Appendix Table~A8). Under A/B imbalance this is a floor diagnostic, not a ranking; VisionZip's paired in-stack ceiling agrees ($0.556$ vs. $0.558$ NDCG).

\section{Verbalization Defines the Image--Text Memory Boundary}
\label{sec:fact-transfer}

Low visual dependence is one route to apparent safety: the future does not need the image. A second is distinctly multimodal---the future may need a visual fact that no longer requires image KV because an earlier assistant answer wrote it into persistent text KV. \CVMA{} next asks when this substitution occurs.

\subsection{A second route to apparent safety}
A $2{\times}2$ image/assistant-output intervention finds cross-source redundancy that does not simply strengthen with every answer (immediate-next-turn interactions $+0.164/+0.049$ then $-0.115$; appendix). VisDial's always-visible caption prevents a clean no-relay control.

\subsection{Verbalization, not mere exposure, defines transfer}
Aggregate redundancy does not reveal what moved. The controlled question is simple: an image contains facts $A$ and $B$; the assistant says $A$ but not $B$; once image KV is unavailable, can a later turn recover $A$, $B$, or both? We build equal-length histories that state one fact and leave the other unstated, then probe them separately; facts pass independent Qwen-32B/InternVL-9B judgment and outcome-blind review before assignment, and 160 pseudowords give matched nonvisual controls. An $S{\times}W{\times}R$ design varies whether the target is stated, whether the answering pass can read the image, and whether the later probe can; every branch is recomputed under its source intervention.

The untouched 80-image confirmation completes 1{,}920/1{,}920 exact-audited cells (Appendix Figure~A1). The target facts genuinely need vision (removing probe image access costs $+1.778$ in the nonce-adjusted correct--foil margin); once a fact is stated, text alone raises that margin by $+3.874$ without image access and reduces the later value of image KV by $+2.215$ ($[1.853,2.579]$), positive for 79/80 image means. Put plainly, saying $A$ makes $A$ recoverable from text once the image is gone.

The stronger hidden-carrier explanation fails: seeing $B$ while saying $A$ leaves no reliable answer-discriminative trace of $B$ (interaction $-0.177$, $[-0.321,-0.035]$; pseudoword-controlled), and this replicates on Idefics (Appendix Table~A8). The reliable boundary is exposure of the particular fact in language, not mere exposure to the image.

\subsection{The boundary survives stock-generated history}
A prospective test stock-generates one 7B description per frozen image, then seals tokens and mention labels before scoring. In 960/960 factorial cells, explicit facts acquire no-image text memory ($+3.336$) and substitution ($+2.898$, $[2.248,3.582]$) with near-zero nonce interaction; mention is model-selected, so the explicit--absent contrast is descriptive while each within-fact source intervention stays causal. The separation replicates on Idefics (explicit substitution $+3.801$ versus absent $+0.258$), so it is not Qwen-specific, though only Qwen shows a stable direct probability uplift (Appendix Table~A8).

\section{Related Work}

Prior work approaches unknown future relevance from three sides. Text-cache methods infer importance from attention or observation windows, learn future-aware retention, and compress long-context or multi-turn caches \citep{li2024snapkv,chen2024nacl,xiao2024efficient,tang2026predicting,bui2026make,chen2026nestedkv,chen2026sonic,liu2025flowkv}; multimodal caches add modality-, layer-, head-, and image-order structure or keep a broad retrievable pool for changing relevance \citep{wan2024look,tu2025vl,wang2026prefixkv,huang2025aircache,zeng2026hybridkv,zhuang2026myopia,khaki2025sparsevila,wang2026rethinking,liu2026retentivekv,chen2026last}; and quantization keeps every address at low precision instead of deleting \citep{liu2024kivi,yang2024no,han2025calibquant,yang2026revisiting}; coverage is thus our matched-storage control, not a contribution. Visual-to-text migration and verbalized memory are also known \citep{lin2025boosting,chatterjee2025memory}. Our object is instead the separating intervention that tests these systems: on one trajectory, paired region drops, random and marginal-utility controls, an all-image drop, and a fact-level source test jointly decide whether a score identifies causal utility, whether selectable signal exists, and whether low dependence hides a wrong deletion (appendix).

\section{Conclusion}

Evidence is bounded (one non-Qwen model, a Qwen-only policy ladder, and teacher-forced NLL as the primary metric; the appendix gives the full scope). Within these bounds, \CVMA{} yields a unified causal account of safe forgetting: current attention misestimates future utility, the eviction failure surfaces only when later turns need discarded evidence, and assistant text substitutes chiefly for facts it explicitly says. Evaluation should therefore condition on visual dependence and unstated facts, and never treat low current attention as sufficient for irreversible deletion.

\bibliography{references}

\end{document}